\documentclass{article}

\usepackage[sglblindworkshop,final]{neurips_2026}

\usepackage[utf8]{inputenc}
\usepackage[T1]{fontenc}
\usepackage{url}
\usepackage{booktabs}
\usepackage{graphicx}
\usepackage{amsmath}
\usepackage{amssymb}
\usepackage{nicefrac}
\usepackage{microtype}
\usepackage{xcolor}

\usepackage{multirow}
\usepackage{tikz}
\usetikzlibrary{arrows.meta, calc, positioning}

\newcommand{\JF}{\mathcal{J}\&\mathcal{F}}
\newcommand{\evpxs}{ev/px/s}

\usepackage{hyperref}

\workshoptitle{Grounded and Faithful Vision-Language Models for Real-World Deployment}

\title{Event-SAMA: Language-Grounded Video Object Segmentation\\
       on Neuromorphic Event Streams}

\author{%
  Sergei Fedorchenko\textsuperscript{*} \quad
  Max Zehnder\textsuperscript{*} \quad
  Nikola Zubić \quad
  Rong Zou \quad
  Davide Scaramuzza \\
  University of Zurich \\
  \texttt{sergey.a.fedorchenko@gmail.com} \\
  \texttt{max.zehnder@uzh.ch} \\
  \textsuperscript{*}Equal contribution.
}

\begin{document}

\maketitle

\begin{abstract}
Event cameras offer microsecond temporal resolution, high dynamic range and no motion blur, but referring video object segmentation (RVOS) lacks event-stream datasets and models.
We introduce \textbf{Event-SAMA-239K}, a synthetic-event conversion of the SAMA-239K grounded video-chat corpus and all additional SAMA training data: 30,650 video clips and 392,162 images.
Optical-flow, FILM or saccade upsampling followed by a linear event generator preserves all mask, referring-expression and dialogue annotations.
We also introduce \textbf{Event-SAMA}, adapting SAMA-4B with a two-channel polarity patch embedding initialised from pretrained RGB weights and a zero-initialised event adapter.
End-to-end training yields \textbf{24.37} $\mathcal{J}\&\mathcal{F}$ on MeViS \texttt{valid\_u} and 22.33 $\mathcal{J}\&\mathcal{F}$ on the converted ReVOS \texttt{valid} subset, establishing the first event-native RVOS baseline (RGB SAMA-4B: 55.4 on MeViS).
Frozen RGB SAMA-4B with pseudo-RGB renderings or E2VID reconstructions of the same events scores 13.23 and 15.94 $\mathcal{J}\&\mathcal{F}$.
Without retraining, Event-SAMA works on real event-camera recordings when the accumulation window matches their event density.
A seed-controlled ablation with a measured noise floor of 0.47 $\mathcal{J}\&\mathcal{F}$ finds gains from diverse training sources over MeViS alone ($+4.99$), RGB-mean event-stem initialisation ($+4.33$), and a trainable SAM2 mask decoder ($+2.76$); the event adapter, partial ViT unfreezing and event polarity are within noise at this budget.
An earlier, partially converted corpus snapshot trained with the identical recipe scores 1.38 $\mathcal{J}\&\mathcal{F}$ higher on MeViS and 1.12 higher on ReVOS.
Both are single runs that also differ in schedule length, so we draw no conclusion about corpus scale.
\end{abstract}

\hfill \break
\noindent \textbf{Multimedial Material:} For code, dataset generation process, weights and more visit:\\ \url{https://github.com/uzh-rpg/Event-SAMA}.

\section{Introduction}
\label{sec:intro}

Referring video object segmentation (RVOS) and grounded video dialogue have
progressed rapidly on RGB video, driven by motion-centric benchmarks such as
MeViS~\citep{ding2023mevis} and by models that couple a large language model
(LLM) to a promptable segmenter~\citep{yuan2025sa2va, sun2025sama}. These
models ground language in pixels, which is exactly what real-world deployment
of vision-language models requires. All of them, however, assume dense,
synchronous RGB frames, and they inherit the failure modes of frame-based
cameras: motion blur under fast motion, saturation in high-dynamic-range
scenes and noise in low light.

Event cameras report per-pixel brightness changes asynchronously with
microsecond latency, a dynamic range above 120\,dB and no motion blur. They
are attractive precisely in the regimes where frame-based cameras degrade, but
the language-grounded segmenters that have matured on RGB video cannot consume
this modality. To our knowledge, no dataset or model addresses referring
segmentation from natural language on event streams. Recording a
language-annotated event corpus at the scale that current vision-language
models require would mean re-capturing hundreds of hours of video with event
hardware \emph{and} re-annotating it with referring expressions and multi-turn
dialogue. We instead follow the standard practice in event
vision~\citep{gehrig2020vid2e} and convert an existing, richly annotated RGB
corpus into synthetic event streams, inheriting its annotations at no
annotation cost.

This paper makes three contributions:
\begin{itemize}
  \item \textbf{Event-SAMA-239K} (Section~\ref{sec:dataset}), an event-camera
        conversion of the SAMA-239K grounded video-chat
        corpus~\citep{sun2025sama} and of the full SAMA training mix:
        30{,}650 video clips and 392{,}162 images from ten source datasets,
        converted with optical-flow, FILM or saccade upsampling followed by a
        linear event generator~\citep{ziegler2023realtime}. All SAMA-239K mask,
        referring-expression and dialogue annotations are preserved.
  \item \textbf{Event-SAMA} (Section~\ref{sec:method}), an event-native
        adaptation of SAMA-4B in which the InternViT patch embedding is
        replaced by a two-channel ON/OFF polarity stem initialised from the
        pretrained RGB weights, and a zero-initialised adapter aligns event
        features with the frozen vision-language pipeline.
  \item A \textbf{seed-controlled ablation study} and an
        \textbf{event-specific evaluation protocol}
        (Section~\ref{sec:experiments}) that separate real effects from
        run-to-run noise and characterise the sensitivity of RVOS on events to
        keyframe count and accumulation-window width.
\end{itemize}

\section{Related work}
\label{sec:related}

\paragraph{Grounded video segmentation and dialogue.}
Ref-YouTube-VOS~\citep{seo2020urvos} and MeViS~\citep{ding2023mevis}
established referring VOS, the latter with a focus on motion-centric
expressions. SAM2~\citep{ravi2024sam2} provides promptable segmentation with
temporal propagation, and Sa2VA~\citep{yuan2025sa2va} couples it to a
multimodal LLM so that language can drive mask prediction through a dedicated
\texttt{[SEG]} token, following the LISA reasoning-segmentation
recipe~\citep{lai2024lisa}. SAMA~\citep{sun2025sama} extends Sa2VA to
multi-turn referential dialogue with a spatio-temporal context aggregator and
contributes the SAMA-239K training corpus and SAMA-Bench. We build directly on
the SAMA-4B model and the SAMA-239K data; our contribution is the event
modality, not the underlying architecture or the source annotations.

\paragraph{Event vision and synthetic events.}
Labelled event data is scarce, so converting RGB video to synthetic event
streams is standard practice~\citep{gehrig2020vid2e}.
\citet{ziegler2023realtime} benchmark several simulators and propose linear
pixel-difference thresholding between frames, which matches the
event density of log-intensity generators at substantially lower cost; we
adopt it as our conversion path. Downstream event research has concentrated
on low-level tasks such as optical flow, reconstruction and detection.
EventGPT~\citep{eventgpt2025} is, to our knowledge, the only prior work
coupling an LLM with event streams; it supports captioning, reasoning and
visual question answering, but produces text only and performs neither
pixel-level segmentation nor temporal object tracking. No prior dataset or
model addresses referring video object segmentation on event data.
A second route to labelled \emph{real} events transfers labels from
co-recorded frames: EV-SegNet~\citep{alonso2019evsegnet} labels DDD17 by
running a segmentation network on the grayscale frames, and
DSEC-Semantic~\citep{sun2022ess} pseudo-labels frames warped into the event
view and filters them by manual inspection. These pipelines yield class labels
for driving scenes with few classes (DSEC-Semantic: 11, daytime only, because
the frame camera fails in hard illumination); referring expressions and
dialogue about individual instances cannot be derived from frames and would
require new human annotation for every clip. We therefore convert an existing
referring corpus for training and use frame-annotated real data
(MouseSIS~\citep{hamann2024mousesis}) for evaluation (Section~\ref{sec:real}).

\section{Event-SAMA-239K}
\label{sec:dataset}

\subsection{Sources and scope}
\label{sec:scope}

SAMA-239K~\citep{sun2025sama} unifies five video datasets (VidSTG, SA-V,
LV-VIS, Ref-YouTube-VOS, MeViS) under a common schema of object masks,
referring expressions, region descriptions and multi-turn grounded
conversations. Beyond this annotated core, the full SAMA training recipe
draws on additional video and image instruction-tuning data (ChatUniVi,
ReVOS, LLaVA-1.5, COCO, GRand-F). We convert both tiers and inherit
all original annotations; the dataset is named after its annotated
SAMA-239K core. Table~\ref{tab:dataset} lists the composition, upsampling method and contrast threshold for event generation per source. The \emph{Count} column reports the annotated clips or images used for training; additional clips from the same sources were converted without mask annotations and may support self-supervised pre-training but are not counted here. The event streams are stored as gzip-compressed HDF5 files with int32 microsecond timestamps and total 17.41\,TB. Licensing and release are described in Appendix~\ref{app:license}.

\begin{table}[t]
  \caption{Datasets converted to synthetic event streams. \emph{Count} is
  video clips for video datasets and images for image datasets (annotated). $c_\pm$ is
  the contrast threshold of the linear event generator. \emph{\evpxs} is the
  median number of events per pixel per second (2\,s cap) over 200 randomly
  sampled items per dataset. $^\dagger$ReVOS comprises TAO (445 clips), OVIS
  (140) and UVO (77).}
  \label{tab:dataset}
  \centering
  \small
  \setlength{\tabcolsep}{5pt}
  \begin{tabular}{lllrrr}
    \toprule
    Dataset & Type & Upsampling & $c_\pm$ & Count & \evpxs \\
    \midrule
    \multicolumn{6}{l}{\textit{Event-SAMA-239K core (masks, referring expressions, dialogue)}} \\
    \midrule
    VidSTG          & Video & optical flow & 3 &   6{,}618 & 26.6 \\
    SA-V            & Video & optical flow & 5 &   2{,}867 & 20.8 \\
    LV-VIS          & Video & optical flow & 3 &   2{,}595 &  9.6 \\
    Ref-YouTube-VOS & Video & optical flow & 5 &   1{,}976 & 55.1 \\
    MeViS           & Video & optical flow & 4 &   1{,}087 &  6.4 \\
    \midrule
    \multicolumn{6}{l}{\textit{Additional SAMA training data}} \\
    \midrule
    ChatUniVi       & Video & FILM         & 3 &  14{,}845 &  4.9 \\
    ReVOS$^\dagger$ & Video & optical flow & 3 &       662 & 29.7 \\
    LLaVA-1.5       & Image & saccade      & 5 & 308{,}379 & 40.8 \\
    COCO            & Image & saccade      & 4 &  82{,}783 & 64.5 \\
    GRand-F         & Image & saccade      & 5 &   1{,}000 & 29.9 \\
    \midrule
    Total           &       &              &   & 422{,}812 & \\
    \bottomrule
  \end{tabular}
\end{table}

\subsection{Conversion pipeline}
\label{sec:pipeline}

Every source is converted with a linear event generator~\citep{ziegler2023realtime}
preceded by one of three upsampling strategies chosen according to content
type. The generator emits an ON event at a pixel whenever its grey value has
risen by $c_+$ compared to the same pixel in the previous frame, and an OFF event whenever it has
fallen by $c_-$; thresholds are tuned per dataset in the range
$c_\pm \in [3, 5]$ (Table~\ref{tab:dataset}).

\paragraph{Video sources (optical flow).}
Consecutive frame pairs are upsampled to at least 200\,fps with DIS optical
flow~\citep{kroeger2016dis}, following the dense interpolation of ~\citet{ziegler2023realtime} before event generation. This is the default path
for all video sources except ChatUniVi.

\paragraph{ChatUniVi (FILM).}
For ChatUniVi we replace flow-based warping with FILM~\citep{reda2022film}, a
learned frame-interpolation model, using the adaptive upsampling scheme of vid2e ~\citep{gehrig2020vid2e}. FILM synthesises intermediate frames from
the perceived motion between adjacent frames and produces smoother luminance
transitions with fewer warping artefacts than optical-flow upsampling.

\paragraph{Image sources (saccade).}
A static image produces no brightness changes and therefore no events. To
obtain a non-empty stream from still images (COCO, LLaVA-1.5, GRand-F) we
apply a synthetic saccade, a $8$\,px shift along three directions over 60
frames spanning 300\,ms, followed by event generation. This mimics the events
a real sensor produces when a static scene is viewed under small camera or eye
movements.

\paragraph{Event representation.}
At training and test time we accumulate events into two-channel polarity
histograms. For a timestamp $t$ and window width $w$, ON and OFF events with
timestamps in $[t - w/2,\; t + w/2)$ are counted into separate channels,
yielding an event frame $E_t \in \mathbb{R}^{2 \times H \times W}$. Counts are
compressed as $\operatorname{asinh}(x)/4$ before entering the network to limit
the dynamic range of dense regions. We use $w = 10$\,ms by default, i.e.\ an
event frame rate of 100\,Hz; the sensitivity to $w$ is analysed in
Appendix~\ref{sec:protocol}.

\subsection{Validation of the converted streams}
\label{sec:validation}

We validate the streams qualitatively and quantitatively. Qualitatively, we
sample ten clips per dataset and inspect the source frames, the per-frame
event accumulation and sub-bin windows within each inter-frame gap
(Appendix~\ref{app:vis}). Quantitatively, we measure the distribution of
events per pixel per second (\evpxs, 2\,s cap) over 200 randomly sampled
items per dataset; medians are reported in Table~\ref{tab:dataset}.

Across the video sources, medians range from 4.9\,\evpxs\ (ChatUniVi) to
55.1\,\evpxs\ (Ref-YouTube-VOS). For reference, \citet{ziegler2023realtime}
report 62.8\,\evpxs\ for a high-dynamic real recording and 0.046\,\evpxs\ for
a low-dynamic one, so our video sources fall within the range of real sensors,
towards its high-dynamic end. Figure~\ref{fig:evdist} compares the \evpxs\
distribution of converted ChatUniVi clips with real Gen1
recordings~\citep{detournemire2020gen1}; the two distributions overlap
substantially (medians 4.9 vs.\ 8.1\,\evpxs). One caveat applies: linear
thresholding also fires on optical-flow warping residuals and JPEG
compression artefacts, so at equal \evpxs\ a real sensor encodes somewhat more
scene motion than our synthetic streams.

Image sources exhibit higher densities (29.9--64.5\,\evpxs) because the
saccade shifts all image content across the contrast threshold
simultaneously, so a large fraction of pixels fires throughout the 300\,ms
window. This is by design, but image-derived streams are therefore denser
than typical real recordings (Appendix~\ref{app:vis},
Figure~\ref{fig:conversion_llava}).

\section{Event-SAMA}
\label{sec:method}

\paragraph{Problem statement.}
Given an event stream $\mathcal{E} = \{(x_i, y_i, t_i, p_i)\}$ with pixel
coordinates $(x_i, y_i)$, timestamps $t_i$ and polarities
$p_i \in \{\mathrm{ON}, \mathrm{OFF}\}$, together with a natural-language
referring expression $q$, the task is to predict a binary mask $M_t$ of the
referred object at every annotated timestamp $t$. No RGB frame is available at
training or test time.

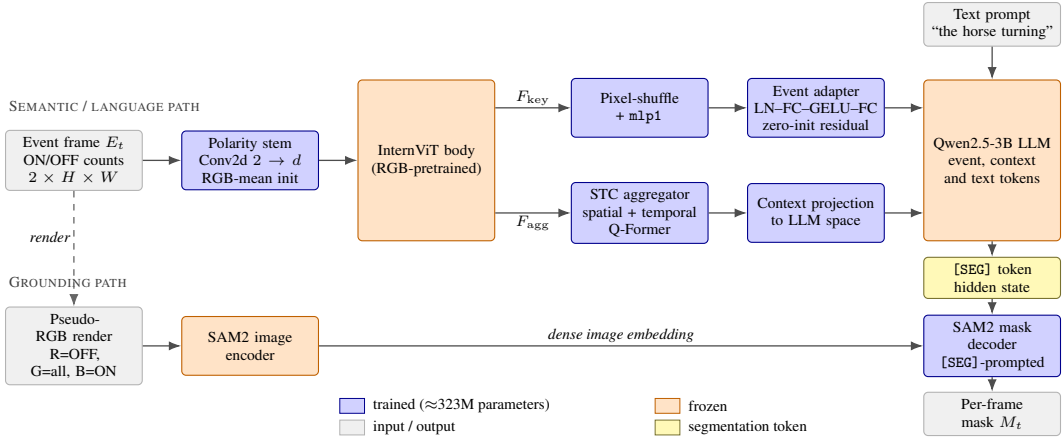
\begin{figure}[t]
  \centering
  \resizebox{\linewidth}{!}{%
  \begin{tikzpicture}[
    font=\scriptsize,
    >=Latex,
    box/.style={draw, rounded corners=2pt, align=center, text width=2.0cm,
                minimum height=0.95cm, inner sep=3pt, line width=0.5pt},
    io/.style={box, fill=gray!12, draw=gray!60},
    trained/.style={box, fill=blue!18, draw=blue!55!black},
    frozen/.style={box, fill=orange!22, draw=orange!75!black},
    segtok/.style={box, fill=yellow!35, draw=olive!70!black, minimum height=0.6cm},
    lbl/.style={font=\scriptsize\itshape, inner sep=1.5pt, fill=white},
    flbl/.style={font=\scriptsize, inner sep=1pt},          
    pathlbl/.style={font=\scriptsize\scshape, text=black!75, inner sep=1.5pt,
                    fill=white, anchor=south west},
    arr/.style={->, line width=0.6pt, draw=black!70},
    darr/.style={->, line width=0.6pt, draw=black!60, dashed},
  ]
    \def\xa{0}\def\xb{2.85}\def\xc{5.7}\def\xd{9.15}\def\xe{12.0}\def\xf{14.85}

    \node[io]      (ev)   at (\xa,0)     {Event frame $E_t$\\ ON/OFF counts\\ $2\times H\times W$};
    \node[trained] (stem) at (\xb,0)     {Polarity stem\\ Conv2d $2\to d$\\ RGB-mean init};
    \node[frozen, minimum height=2.6cm] (vit) at (\xc,0) {InternViT body\\ (RGB-pretrained)};
    \node[trained] (mlp)  at (\xd,0.85)  {Pixel-shuffle\\ + \texttt{mlp1}};
    \node[trained] (ad)   at (\xe,0.85)  {Event adapter\\ LN--FC--GELU--FC\\ zero-init residual};
    \node[trained] (stc)  at (\xd,-0.85) {STC aggregator\\ spatial + temporal\\ Q-Former};
    \node[trained] (proj) at (\xe,-0.85) {Context projection\\ to LLM space};
    \node[frozen, minimum height=2.6cm] (llm) at (\xf,0) {Qwen2.5-3B LLM\\ event, context\\ and text tokens};
    \node[io, minimum height=0.7cm] (txt) at (\xf,2.2) {Text prompt\\ ``the horse turning''};
    \node[segtok]  (seg)  at (\xf,-1.9)  {\texttt{[SEG]} token\\ hidden state};
    \node[trained] (dec)  at (\xf,-3.0)  {SAM2 mask decoder\\ \texttt{[SEG]}-prompted};
    \node[io, minimum height=0.7cm] (out) at (\xf,-4.1) {Per-frame mask $M_t$};

    \node[io]      (prgb) at (\xa,-3.0) {Pseudo-RGB render\\ R=OFF, G=all, B=ON};
    \node[frozen]  (sam)  at (\xb,-3.0) {SAM2 image\\ encoder};

    \node[pathlbl] at ($(ev.north west)+(0,0.25)$)   {Semantic / language path};
    \node[pathlbl] at ($(prgb.north west)+(0,0.25)$) {Grounding path};

    \draw[arr] (ev) -- (stem);
    \draw[arr] (stem) -- (vit);
    \draw[arr] (vit.east |- mlp) -- node[flbl, above] {$F_{\mathrm{key}}$} (mlp);
    \draw[arr] (mlp) -- (ad);
    \draw[arr] (ad) -- (llm.west |- ad);
    \draw[arr] (vit.east |- stc) -- node[flbl, below] {$F_{\mathrm{agg}}$} (stc);
    \draw[arr] (stc) -- (proj);
    \draw[arr] (proj) -- (llm.west |- proj);
    \draw[arr] (txt) -- (llm);
    \draw[arr] (llm) -- (seg);
    \draw[arr] (seg) -- (dec);
    \draw[arr] (dec) -- (out);
    \draw[darr] (ev.south) -- node[lbl, left, pos=0.4] {render} (prgb.north);
    \draw[arr] (prgb) -- (sam);
    \draw[arr] (sam) -- node[lbl, above] {dense image embedding} (dec);

    \begin{scope}[every node/.style={font=\scriptsize, anchor=west, inner sep=1pt}]
      \fill[blue!18, draw=blue!55!black]     (4.3,-3.85) rectangle (4.7,-4.09);
      \node at (4.8,-3.97) {trained ($\approx$323M parameters)};
      \fill[gray!12, draw=gray!60]           (4.3,-4.20) rectangle (4.7,-4.44);
      \node at (4.8,-4.32) {input / output};
      \fill[orange!22, draw=orange!75!black] (9.4,-3.85) rectangle (9.8,-4.09);
      \node at (9.9,-3.97) {frozen};
      \fill[yellow!35, draw=olive!70!black]  (9.4,-4.20) rectangle (9.8,-4.44);
      \node at (9.9,-4.32) {segmentation token};
    \end{scope}
  \end{tikzpicture}%
  }
  \caption{Event-SAMA, adapted from SAMA-4B. Each event frame $E_t$ drives two
  paths. On the \emph{semantic/language path}, the two-channel polarity stem
  (trained) feeds the frozen InternViT; per-keyframe features
  $F_{\mathrm{key}}$ pass through the pixel-shuffle/\texttt{mlp1}
  projection and the trained event adapter, while whole-clip context
  $F_{\mathrm{agg}}$ is produced by the trained STC aggregator. Both are
  interleaved with the text prompt and fed to the frozen Qwen2.5-3B LLM,
  which emits a \texttt{[SEG]} token. On the \emph{grounding path}, $E_t$ is
  rendered to pseudo-RGB for the frozen SAM2 image encoder; the trained SAM2
  mask decoder, prompted by the \texttt{[SEG]} hidden state, outputs the
  per-frame mask.}
  \label{fig:arch}
\end{figure}

Event-SAMA adapts SAMA-4B~\citep{sun2025sama}, which itself builds on
Sa2VA~\citep{yuan2025sa2va}, SAM2~\citep{ravi2024sam2} and the
BLIP-2~\citep{li2023blip2} Q-Former, to consume the event frames $E_t$ of
Section~\ref{sec:pipeline} in place of RGB images. Figure~\ref{fig:arch}
shows the architecture and marks the trained components. We first summarise
the unchanged SAMA-4B pipeline and then describe the three event-specific
modifications.

\paragraph{Base pipeline (unchanged from SAMA-4B).}
Each sampled keyframe is patch-embedded and encoded by a frozen
InternViT~\citep{chen2024internvl}, pixel-shuffled to reduce the token count
and linearly projected by \texttt{mlp1} into the language embedding space,
giving per-keyframe features $F_{\mathrm{key}}$. In parallel, a two-stage
spatio-temporal context (STC) aggregator built on the BLIP-2
Q-Former~\citep{li2023blip2} summarises the whole clip into a fixed-size
context $F_{\mathrm{agg}}$ (details in Appendix~\ref{app:impl}). Both are
interleaved with the text tokens and fed to a frozen
Qwen2.5-3B~\citep{qwen25} LLM. Following LISA~\citep{lai2024lisa}, the
vocabulary contains a \texttt{[SEG]} token; when the LLM emits it, the token's
final hidden state serves as a sparse prompt embedding for the SAM2 mask
decoder, analogous to a point or box prompt. The frozen SAM2 image encoder
supplies the dense image embedding, and the mask decoder, trained in both
SAMA-4B and Event-SAMA, produces the per-frame mask.

\paragraph{Two-channel polarity stem.}
The pretrained patch embedding is a $\mathrm{Conv2d}(3 \to d)$ over RGB. We
replace it with a $\mathrm{Conv2d}(2 \to d)$ over the ON/OFF channels of
$E_t$. Its kernel is initialised by averaging the pretrained RGB kernel over
the input dimension and broadcasting the result to both event channels; the
bias is copied unchanged. The pretrained ViT body therefore receives inputs
in a familiar range from the first iteration instead of having to adapt to a
randomly initialised stem. This initialisation is worth $+4.33$ $\JF$ over a
random one (Table~\ref{tab:ablation}). Everything downstream of the patch
embedding within InternViT is untouched.

\paragraph{Event adapter.}
Following the event-language adapter of EventGPT~\citep{eventgpt2025}, we
insert a residual module (LayerNorm $\to$ Linear $\to$ GELU $\to$ Linear,
hidden size unchanged, 8.4M parameters) after \texttt{mlp1} on the keyframe
path. The second linear layer is zero-initialised, so the adapter is the
identity at initialisation: training starts from the exact pretrained
RGB--language alignment, and the adapter can learn a modality-specific
correction on top of it. The aggregated path $F_{\mathrm{agg}}$ is not
adapted because the STC aggregator and its output projection are trained from
scratch, as in the base SAMA recipe. In the ablation, removing the adapter
changes the score by $+0.67$ $\JF$, and on an independent earlier batch by
$-0.57$; both are within the seed noise (Section~\ref{sec:ablations}). We
retain the adapter on architectural rather than empirical grounds:
zero-initialisation guarantees that it cannot hurt at the start of training,
and its cost is negligible.

\paragraph{SAM2 grounding path.}
The SAM2 image encoder expects three-channel input. Rather than retraining it
on two channels, we render each event frame to a pseudo-RGB image
(R = OFF density, G = total density, B = ON density) and apply SAM2's standard
preprocessing. This reuses the RGB-pretrained encoder unmodified at the cost
of a synthetic-image domain gap (Section~\ref{sec:limitations}). Unlike the
vision-to-language path, the mask decoder is fully trained: its output space
(per-pixel mask logits) is task-specific, so there is no pretrained alignment
to preserve. Freezing the decoder costs $2.76$ $\JF$
(Table~\ref{tab:ablation}).

\paragraph{Trainable components and training.}
The LLM, the InternViT body and the SAM2 image encoder are frozen. The trained
components are the polarity stem (0.4M parameters), the event adapter (8.4M),
the \texttt{mlp1} vision--language projection that the adapter wraps (12.6M),
the STC aggregator with its projections (288.8M), the context projection into
the LLM embedding space (4.2M), the projection mapping \texttt{[SEG]} hidden
states into the SAM2 prompt space (4.7M), and the SAM2 mask decoder (4.2M),
for a total of 323M trained parameters. The \texttt{[SEG]} token requires no
new embedding: it is already part of the SAMA-4B vocabulary, so its input
embedding is inherited from SAMA-4B training and remains frozen with the rest
of the LLM. The trained \texttt{[SEG]}-to-prompt projection is the only
component that adapts its representation.

An \emph{item} is one annotated training sample, i.e.\ an expression--clip
pair, a referring-expression image sample, or one dialogue. An \emph{epoch} is
one pass over the repeat-weighted concatenation of all sources, with
per-source repeat factors of 1--10 up-weighting the small mask-annotated
sources (e.g.\ MeViS $\times 4$). One epoch of the full corpus is
${\approx}1.60$M weighted samples, so 4 epochs at global batch 48 give
133{,}208 iterations.

\section{Experiments}
\label{sec:experiments}

\subsection{Setup}
\label{sec:setup}

\paragraph{Benchmarks and metrics.}
We evaluate referring VOS on MeViS \texttt{valid\_u} (50 videos, 793
video--expression pairs) and on ReVOS \texttt{valid}~\citep{sun2025sama}
(416 videos), reporting region similarity $\mathcal{J}$, contour accuracy
$\mathcal{F}$ and their mean $\JF$. MeViS \texttt{valid\_u} is fully
converted. ReVOS coverage in our conversion is partial: 97 of the 416
validation videos have a converted event stream. We therefore score ReVOS on
this 97-video subset only, since scoring the unconverted videos as misses
would measure conversion coverage rather than model quality.

\paragraph{Evaluation protocol.}
Because event streams are continuous, a frame-level protocol must be defined
explicitly. We predict masks on $K = 32$ temporally uniform keyframes per
clip, each accumulated over a 10\,ms window centred on the keyframe timestamp
(matching training), and propagate each prediction to the remaining annotated
timestamps by nearest-keyframe assignment before scoring against the
per-frame ground truth. The LLM receives the first 5 keyframes as visual
context. We evaluate the final training checkpoint directly. All numbers in
this paper use this identical setting; its sensitivity is characterised in
Appendix~\ref{sec:protocol}.

\subsection{Main results}
\label{sec:main}

\begin{table}[t]
  \caption{Referring VOS on MeViS \texttt{valid\_u} and ReVOS
  \texttt{valid}$^\ddagger$. Both Event-SAMA rows use the same 4-epoch recipe
  and differ only in training data. The full-corpus model (bold) is our main
  result because it is the model trained on the corpus described in Section~\ref{sec:dataset}. The RGB
  SAMA-4B number is the published MeViS result~\citep{sun2025sama}; it serves
  as an upper bound, not as a like-for-like comparison. The two
  reconstruct-and-reuse rows feed pseudo-RGB renderings or
  E2VID~\citep{rebecq2019e2vid} reconstructions of the same events to the
  unmodified RGB SAMA-4B with its own evaluation script.}
  \label{tab:main}
  \centering

  \footnotesize
  \setlength{\tabcolsep}{3pt}
  \renewcommand{\arraystretch}{0.95}

  \begin{tabular}{@{}llcccc@{}}
    \toprule
    Model & Training data
      & $\mathcal{J}$ & $\mathcal{F}$
      & MeViS $\JF$ & ReVOS $\JF^\ddagger$ \\
    \midrule
    SAMA-4B (RGB, upper bound)
      & SAMA mix (RGB)
      & --- & --- & 55.4 & --- \\
    \midrule
    SAMA-4B (frozen) & pseudo-RGB events & 12.42 & 14.03 & 13.23 & --- \\
    SAMA-4B (frozen) & E2VID reconstr. & 15.62 & 16.25 & 15.94 & --- \\
    \midrule
    Event-SAMA
      & 5-source subset (27{,}461)
      & 25.32 & 26.19 & 25.75 & 23.45 \\
    \textbf{Event-SAMA}
      & \textbf{Full corpus (422{,}812)}
      & \textbf{23.81} & \textbf{24.92}
      & \textbf{24.37} & \textbf{22.33} \\
    \bottomrule
  \end{tabular}

  \vspace{1pt}
  \parbox{0.96\columnwidth}{%
    \scriptsize
    $^\ddagger$Computed on the 97/416 \texttt{valid} videos with a
    converted event stream (Section~\ref{sec:setup}).%
  }
\end{table}

Table~\ref{tab:main} reports the main results. Event-SAMA trained on the full
corpus reaches 24.37 $\JF$ on MeViS \texttt{valid\_u} and 22.33 $\JF$ on the
converted ReVOS subset. The model trained on the five-source subset scores
about one point higher on both benchmarks (25.75 and 23.45). We nevertheless
report the full-corpus model as the main result because it is the model
trained on the corpus this paper describes; selecting the subset
model as the headline because of its higher score would misrepresent the
paper's central model. Section~\ref{sec:ablations} analyses the difference.

A large gap to the RGB upper bound (55.4 $\JF$) remains. Event streams carry
no colour and no static texture, all training and test events are synthetic,
and Event-SAMA is a first baseline rather than a tuned system. A qualitative
example is shown in Appendix~\ref{app:qual}.

\paragraph{Native events vs.\ reconstruct-and-reuse.}
An obvious alternative to an event-native model is to turn events into
RGB-like frames and reuse an existing RGB model. We test two such pipelines
with the frozen RGB SAMA-4B, the checkpoint Event-SAMA is initialised from:
\emph{pseudo-RGB}, the exact rendering our grounding branch feeds to SAM2
(Appendix~\ref{app:impl}), and \emph{E2VID}~\citep{rebecq2019e2vid}, a learned
intensity reconstruction (pretrained, default settings). Both trail the native
model by a wide margin (13.23 and 15.94 vs.\ 24.37 $\JF$; Table~\ref{tab:main}).
Our synthetic events arrive in equal-timestamp bursts between low-rate source
frames, which lies outside E2VID's training distribution and blurs its
reconstructions.

\subsection{Real event-camera data}
\label{sec:real}

All other results use synthetic events. To test whether the model transfers to
real sensors, we evaluate the unchanged full-corpus checkpoint on two public
real-event datasets with ground-truth masks. MouseSIS~\citep{hamann2024mousesis}
provides real $1280{\times}720$ recordings of freely moving mice with per-frame
instance masks; we use all 25 annotated clips, none of which we train on, with
the expression \emph{``the mice''} and the absent \emph{``a red car''} as a
negative control. N-Caltech101~\citep{orchard2015ncaltech} provides real ATIS
recordings with object contours. Setup details and qualitative examples are in
Appendix~\ref{app:real}.

On the 25 MouseSIS clips, \emph{``the mice''} reaches 31.7 $\JF$, against 24.5
for a language-free motion mask computed from the same events, and
\emph{``a red car''} yields an empty mask in 63\% of frames. Since MouseSIS
events are much sparser than our training data, we match the accumulation
window to the event density of each clip instead of using the 10\,ms training
window. On N-Caltech101 the correct class expression reaches a higher IoU than
a wrong-class expression in 69\% of samples, with a mean IoU of 0.52 vs.\ 0.37.
Figure~\ref{fig:mousesis} in Appendix~\ref{app:real} shows qualitative examples.

\subsection{Ablations}
\label{sec:ablations}

\paragraph{Protocol.}
The ablation study was conducted before the full-corpus conversion was
complete and therefore uses the five-source subset (27{,}461 items) with a
1-epoch iso-compute recipe. The baseline is trained with two seeds
(24.20 and 24.67 $\JF$, spread 0.47), which bounds the run-to-run noise; each
variant is trained once. Architectural findings (adapter, decoder
trainability, stem initialisation, polarity, ViT unfreezing) are model
properties and are unaffected by the choice of subset, but the baseline
(24.44) is not the same model as the full-corpus headline (24.37).
Table~\ref{tab:ablation} groups the variants by whether their effect exceeds
the noise floor.

\begin{table}[t]
  \caption{Ablations on MeViS \texttt{valid\_u}, 1-epoch iso-compute recipe
  on the five-source subset. The baseline is the mean of two seeds
  (24.20 / 24.67 $\JF$; spread 0.47). Rows in the first group lie outside the
  noise floor; rows in the second group do not, and the adapter and ViT
  variants reversed sign ($-0.57$, $+0.26$) on an independent earlier batch,
  whereas the decoder effect kept its sign in both batches. The last row is
  not iso-compute with the others (see footnote).}
  \label{tab:ablation}
  \centering
  \small
  \begin{tabular}{lccc}
    \toprule
    Configuration & $\mathcal{J}$ & $\mathcal{F}$ & $\Delta\,\JF$ \\
    \midrule
    Full model (baseline, mean of 2 seeds) & 23.42 & 25.46 & --- \\
    \midrule
    \multicolumn{4}{l}{\textit{Effects outside the noise floor}} \\
    $-$ diverse data mix (train on MeViS only)     & 18.19 & 20.71 & $-4.99$ \\
    $-$ RGB-mean stem init (random init)$^\dagger$ & 16.79 & 20.63 & $-4.33$ \\
    $-$ trainable SAM2 decoder (frozen)            & 20.65 & 22.70 & $-2.76$ \\
    \midrule
    \multicolumn{4}{l}{\textit{Effects within the noise floor}} \\
    $-$ event polarity (ON/OFF merged)             & 24.41 & 25.51 & $+0.52$ \\
    $-$ event adapter                              & 24.26 & 25.95 & $+0.67$ \\
    $+$ ViT last 6 layers unfrozen                 & 22.89 & 24.58 & $-0.71$ \\
    \midrule
    \multicolumn{4}{l}{\textit{Data scale (4-epoch recipe, not iso-compute)}} \\
    $+$ full corpus (422{,}812 items)$^\ddagger$   & 23.81 & 24.92 & $-1.38$ \\
    \bottomrule
  \end{tabular}
  \\[2pt]
  \footnotesize$^\dagger$Trained in an earlier batch against a matched
  baseline of 23.04 $\JF$; the effect size makes a re-run unnecessary.
  $^\ddagger$Delta with respect to the matched 4-epoch five-source model
  (25.75 $\JF$, Table~\ref{tab:main}), not the 1-epoch baseline above.
\end{table}

\paragraph{Data diversity is the dominant factor, up to a point.}
The full-corpus model scores 1.38 $\mathcal{J}\&\mathcal{F}$ below the
comparison model on MeViS and 1.12 below on ReVOS. The two runs differ in
conversion snapshot (${\approx}4\%$ of weighted samples), in schedule length
(127{,}468 vs.\ 133{,}208 iterations under the same 4-epoch cosine schedule),
and in seed, and each was trained once. Against the 0.47
$\mathcal{J}\&\mathcal{F}$ two-seed spread measured for the 1-epoch recipe, we
read this gap as run-to-run variation and draw no conclusion about
training-set size from this pair.

\paragraph{Stem initialisation and decoder trainability.}
Initialising the two-channel stem from the RGB-mean kernel rather than
randomly is worth $4.33$ $\JF$, and training the SAM2 mask decoder rather than
freezing it is worth $2.76$ $\JF$. Both effects are several times larger than
the noise floor, and the decoder effect kept its sign across two independent
training batches.

\paragraph{Adapter and ViT unfreezing.}
Removing the event adapter ($+0.67$) and unfreezing the last six ViT layers
($-0.71$) both produce deltas comparable to the seed spread, and both reversed
sign on an earlier batch ($-0.57$ and $+0.26$). At this budget neither can be
distinguished from a no-op.

\paragraph{Polarity: no measurable effect, but underpowered.}
Merging the ON and OFF channels into their mean, which destroys polarity while
preserving total event count, changes the score by $+0.52$ $\JF$, within the
seed spread. We do \emph{not} interpret this as evidence that polarity is
unnecessary: a single run against a two-seed baseline cannot establish the
absence of an effect of this size, and unlike the adapter and ViT variants
this configuration was measured only once. The supported conclusion is
narrower: at this training budget, polarity is not among the factors that
dominate performance, in contrast to data mix, stem initialisation and decoder
trainability. Whether polarity-free representations are viable for
language-grounded segmentation, which would matter for sensor bandwidth and
representation design, requires a properly powered study.

\section{Limitations}
\label{sec:limitations}

\paragraph{Synthetic events.}
All events are simulated from RGB video with a linear generator. Our
validation (Section~\ref{sec:validation}) matches event \emph{density}
against real Gen1 recordings, but density is only one axis: we do not
reproduce sensor noise, refractory periods, per-pixel threshold mismatch or
the true microsecond timing statistics of a real sensor. Our real-event
evaluation (Section~\ref{sec:real}) is small (25 MouseSIS clips of a single
animal domain and 200 N-Caltech101 samples) and shows that the accumulation
window must be matched to the sensor's event density; training on real or
density-randomised events is the most important direction for follow-up work.

\paragraph{Saccades are a proxy.}
Still-image sources constitute the majority of the corpus by count, and their
events stem from a synthetic saccade rather than scene motion. This is a
deliberate stand-in for eye-movement-driven events and yields higher event
densities (29.9--64.5\,\evpxs) than the video sources; conclusions drawn from
image-heavy training should be read with this in mind.

\paragraph{Partial second benchmark.}
Our best-supported numbers are on MeViS \texttt{valid\_u} (full coverage,
793 items). Our ReVOS \texttt{valid} numbers cover 97 of 416 videos because
the remainder has not yet been converted. The grounded-dialogue capabilities
supported by the training data are trained but not benchmarked.

\paragraph{Ablation budget.}
Ablations use a 1-epoch schedule with two seeds for the baseline and one seed
per variant. Effects below approximately 0.5 $\JF$ cannot be resolved, which
is why three of the six architectural ablations are reported as inconclusive.

\paragraph{Absolute performance.}
At 24.37 $\JF$ against an RGB upper bound of 55.4, Event-SAMA is a first
baseline demonstrating that language-grounded segmentation on events is
feasible, not a competitive system.

\section{Conclusion}
\label{sec:conclusion}

We introduced Event-SAMA-239K, the first large-scale language-grounded event
video dataset (30{,}650 video clips and 392{,}162 images converted to
synthetic event streams with all SAMA-239K annotations preserved), and
Event-SAMA, the first model that segments objects in event streams from
natural-language referring expressions. Event-SAMA reaches 24.37 $\JF$ on
MeViS \texttt{valid\_u} and 22.33 $\JF$ on the converted subset of ReVOS
\texttt{valid}. A seed-controlled ablation shows that data diversity
($+4.99$), RGB-mean initialisation of the event stem ($+4.33$) and a
trainable SAM2 mask decoder ($+2.76$) are the decisive design choices, while
the event adapter, ViT unfreezing and event polarity cannot be separated from
a 0.47-$\JF$ noise floor. Longer accumulation windows at test time improve accuracy even though the
model was trained at 10\,ms. The event-native model outperforms
reconstruct-and-reuse pipelines built on the same RGB model on MeViS and works
on real event-camera recordings without retraining once the accumulation window
matches their event density. We release the conversion pipeline with per-dataset configuration, which regenerate the full corpus from the original RGB source, together with training code and checkpoints, so that language-grounded event
understanding can be studied as a benchmark task.

\begin{ack}
This work was supported by the European Union’s Horizon Europe Research and Innovation Programme under grant agreement No. 101120732 (AUTOASSESS) and the European Research Council (ERC) under grant agreement No. 864042 (AGILEFLIGHT).
\end{ack}

\bibliographystyle{plainnat}
\bibliography{references}

\appendix

\section{Licensing and release}
\label{app:license}
We release the means to regenerate the full corpus rather than the event streams themselves. At 17.41\,TB, the streams are much larger than their RGB sources, so conversion is faster than downloading; several sources, notably SA-V and YouTube-derived corpora, also restrict redistribution of derived works. The release comprises:
\begin{enumerate}
  \item The conversion pipeline, training code, evaluation harness and trained
        checkpoints, which are released under Apache~2.0, matching the SAMA codebase, except the FILM upsampling module adapted from vid2e, which remains under its original GPL-3.0 license.
  \item A complete manifest (conversion method,
        contrast thresholds, upsampling settings) is
        released alongside, so that the full corpus can be regenerated
        by a user who has obtained the original data from its provider.
\end{enumerate}

\section{Implementation details}
\label{app:impl}

\paragraph{Spatio-temporal context aggregator.}
The STC aggregator of SAMA~\citep{sun2025sama} consists of two Q-Formers
built on the BLIP-2 design~\citep{li2023blip2}. A \emph{spatial} Q-Former
with 32 learned query tokens attends over each frame's InternViT features
independently. A \emph{temporal} Q-Former with 2 transformer layers then
attends across the per-frame spatial-Q-Former outputs and produces a
fixed-size clip summary $F_{\mathrm{agg}}$, which
\texttt{Qformer\_final\_proj} maps into the language embedding space. Both
Q-Formers and the projection are trained from scratch, as in the base SAMA
recipe, and operate on event-derived ViT features without further
modification.

\paragraph{Event adapter placement.}
The adapter wraps \texttt{mlp1} in a two-stage sequential module
(\texttt{mlp1} followed by the residual adapter), so that it is applied on
every forward pass regardless of which code path is active at inference time.
\texttt{mlp1} is trained jointly with the adapter (Section~4).

\paragraph{Pseudo-RGB rendering.}
For the SAM2 image encoder, the event frame $E_t$ is rendered to three
channels as $R = E_t^{\mathrm{OFF}}$, $G = E_t^{\mathrm{ON}} + E_t^{\mathrm{OFF}}$,
$B = E_t^{\mathrm{ON}}$, followed by SAM2's standard resizing and
normalisation. The rendering is used only on the grounding path; the
semantic/language path consumes the two-channel frame directly.

\paragraph{Training configuration.}
Training uses 16 NVIDIA GH200 GPUs (4 nodes $\times$ 4 GPUs) with DeepSpeed
ZeRO-2 in bf16, a per-device batch size of 3 (global batch 48) and a cosine
learning-rate schedule with warm-up. The main model is trained for 4 epochs
(133{,}208 iterations) on the full corpus; the five-source comparison model
for 4 epochs on the subset. Ablations use a 1-epoch
schedule on the five-source subset with otherwise identical settings.

\paragraph{Evaluation details.}
For each clip, $K = 32$ keyframe timestamps are placed uniformly over the
annotated duration. Each keyframe is accumulated over a 10\,ms window centred
on its timestamp. The LLM receives the first 5 keyframes as visual tokens;
masks are decoded for all 32 keyframes. Every annotated ground-truth frame is
assigned the prediction of its temporally nearest keyframe before $\mathcal{J}$
and $\mathcal{F}$ are computed with the standard MeViS evaluation code.

\section{Qualitative results}
\label{app:qual}

\begin{figure}[htbp]
  \centering
  \includegraphics[width=\linewidth]{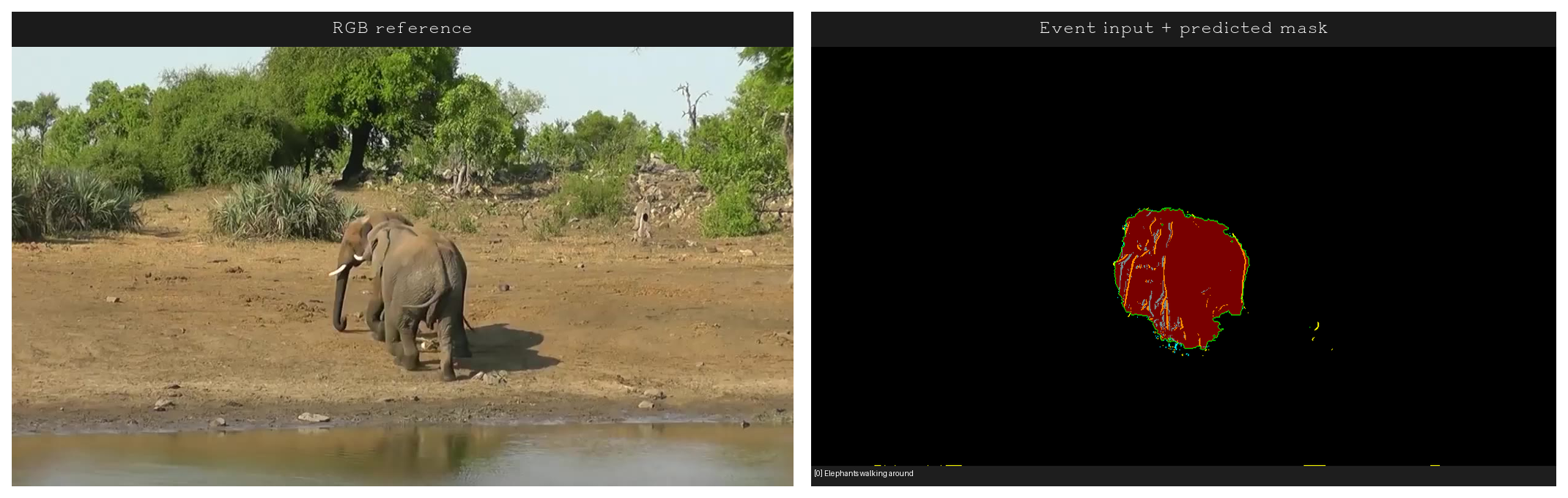}
  \caption{Event-SAMA on MeViS \texttt{valid\_u} for the expression
  ``elephants walking around''. \emph{Left:} the RGB source frame, shown for
  reference only; the model never sees it. \emph{Right:} the two-channel event
  frame the model receives, with the predicted mask overlaid. Only the moving
  elephant generates events; the static background is almost empty.}
  \label{fig:qualitative}
\end{figure}

Figure~\ref{fig:qualitative} illustrates the input the model operates on. The
event frame is extremely sparse compared with an RGB frame: the referred
object is delineated by motion-induced events along its contours, while the
static background (trees, water, ground texture) produces almost no events.
This is the central trade-off of the modality. The model cannot rely on
colour or texture to identify an ``elephant'' and must ground the expression
in shape and motion alone; conversely, a moving object is cleanly separated
from a background that contributes no competing signal.

\section{Protocol sensitivity}
\label{sec:protocol}

\begin{table}[htbp]
  \caption{Protocol sweeps on a fixed checkpoint (MeViS \texttt{valid\_u},
  $\JF$). Left: number of keyframes $K$ at a 10\,ms accumulation window.
  Right: accumulation window $w$ at $K = 32$. Accuracy saturates in $K$ but is
  strongly asymmetric in $w$.}
  \label{tab:protocol}
  \centering
  \small
  \begin{tabular}{lccc@{\hskip 2.5em}lccc}
    \toprule
    Keyframes $K$ & 8 & 16 & 32 & Window $w$ & 5\,ms & 10\,ms & 20\,ms \\
    \midrule
    $\JF$ & 20.64 & 21.93 & 23.04 & $\JF$ & 18.73 & 23.04 & 24.59 \\
    \bottomrule
  \end{tabular}
\end{table}

Table~\ref{tab:protocol} characterises the two protocol parameters on a fixed
checkpoint. Accuracy increases with the number of keyframes with diminishing
returns (20.64 $\to$ 21.93 $\to$ 23.04 $\JF$ for $K = 8, 16, 32$), which
justifies $K = 32$; $K = 64$ could not be evaluated because the SAM2 image
encoder does not run at that batch size in our setup. The accumulation window
shows a strong asymmetry: halving it to 5\,ms costs $4.31$ $\JF$, whereas
doubling it to 20\,ms \emph{gains} $1.55$ $\JF$ even though the model was
trained at 10\,ms. Under-populated event frames are far more damaging than
over-integrated ones, which suggests that future models should be trained at
longer accumulation windows.

\section{Real event-camera evaluation: details}
\label{app:real}

\paragraph{MouseSIS.}
MouseSIS~\citep{hamann2024mousesis} provides real $1280{\times}720$ event
recordings of freely moving mice (Prophesee Gen4.1) with per-frame instance
masks annotated on grayscale frames co-recorded through a beam splitter. We use
all 25 clips with public annotations (the 4 validation and 21 training clips;
15--25\,s each; every third annotated frame, 4{,}946 frames in total); none of
them is used for training. Each clip is queried with \emph{``the mice''}
(ground truth: union of all mice) and with \emph{``a red car''} as a negative
control (object absent). Hot
pixels are removed by a rule set from event statistics alone (isolated pixels
with more than 100 events and more than $10\times$ the median of their
$3{\times}3$ neighbourhood).

\begin{figure}[htbp]
  \centering
  \includegraphics[width=\linewidth]{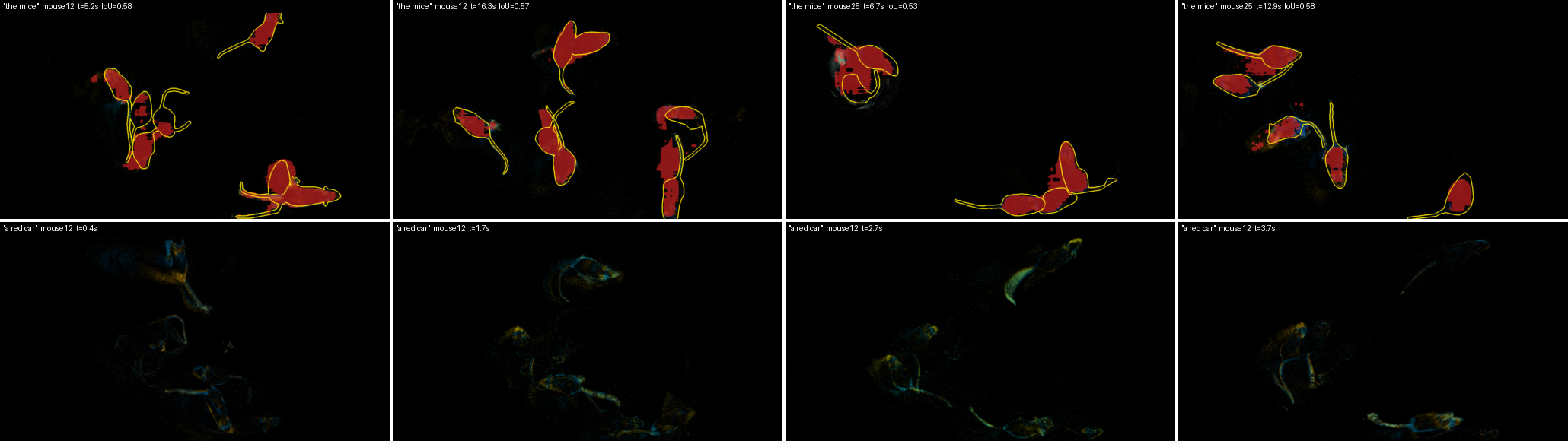}
  \caption{Event-SAMA on real MouseSIS events (75\,ms window), selected
  examples from two validation clips. Background: the events exactly as the
  grounding branch receives them; red: predicted mask; yellow: ground-truth
  contour. Top: \emph{``the mice''}; bottom: the absent \emph{``a red car''}
  (empty predictions). Aggregate numbers are in Section~\ref{sec:real}.}
  \label{fig:mousesis}
\end{figure}

\paragraph{N-Caltech101.}
N-Caltech101~\citep{orchard2015ncaltech} contains real ATIS recordings of
static images under sensor motion. On 200 samples from 25 animal classes,
\emph{``the $\langle$class$\rangle$''} reaches a mean IoU of 0.52 against the
object contour (0.57 at 20\,ms), against 0.37 for a wrong-class expression; the
correct expression scores higher in 69\% of samples.

\paragraph{What differs from synthetic events.}
Three effects absent from our synthetic data matter on real sensors. Event
density shifts the useful window. Hot pixels and background noise distort the
per-window max normalisation of the pseudo-RGB grounding input. Animals that
stop moving emit almost no events and disappear from short windows.

\section{Additional visualisations of the converted data}
\label{app:vis}
\begin{figure}[htbp]
  \centering
  \includegraphics[width=0.8\linewidth]{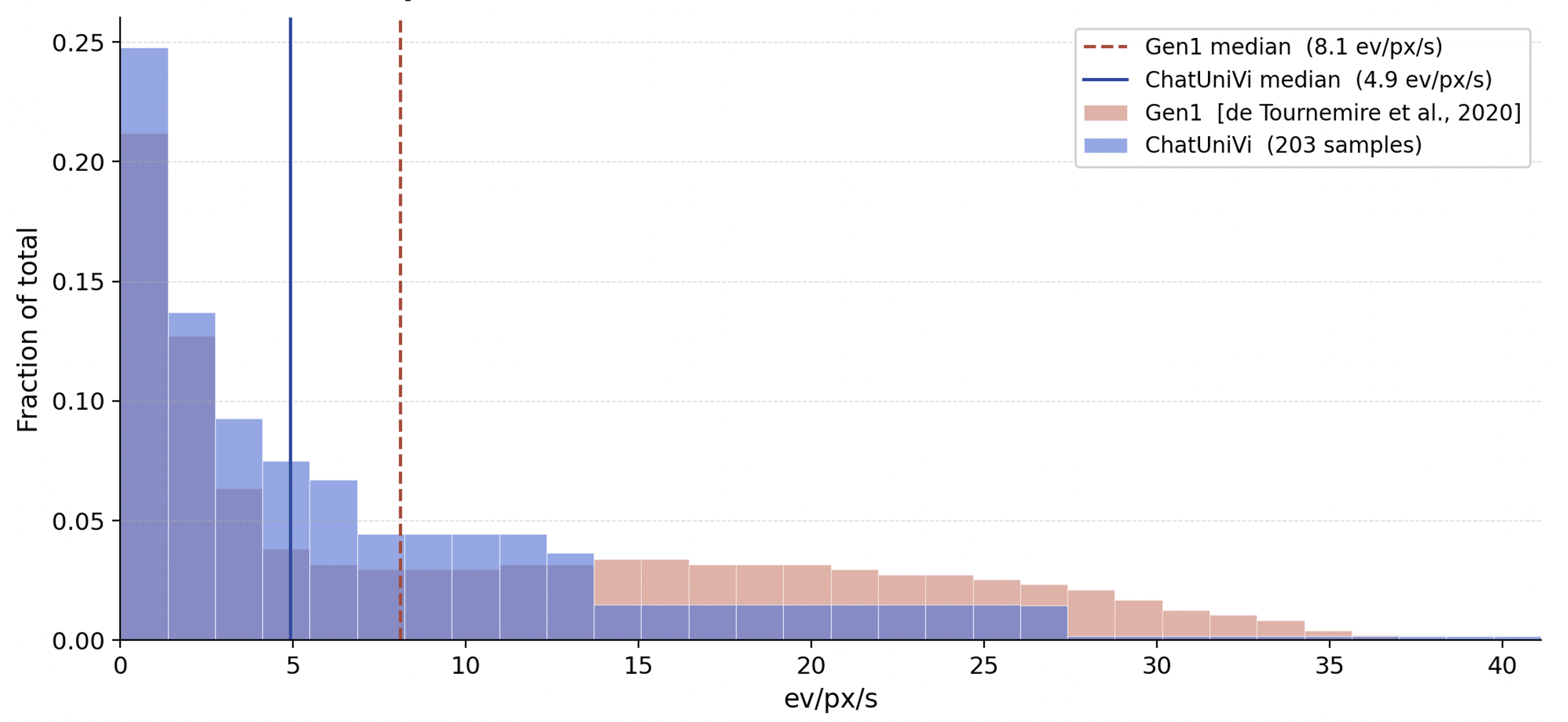}
  \caption{Distribution of events per pixel per second for converted
  ChatUniVi clips (200 clips) and for real Gen1 event-camera
  recordings~\citep{detournemire2020gen1}. Medians: ChatUniVi 4.9,
  Gen1 8.1\,\evpxs.}
  \label{fig:evdist}
\end{figure}

\begin{figure}[htbp]
  \centering
  \includegraphics[width=\linewidth]{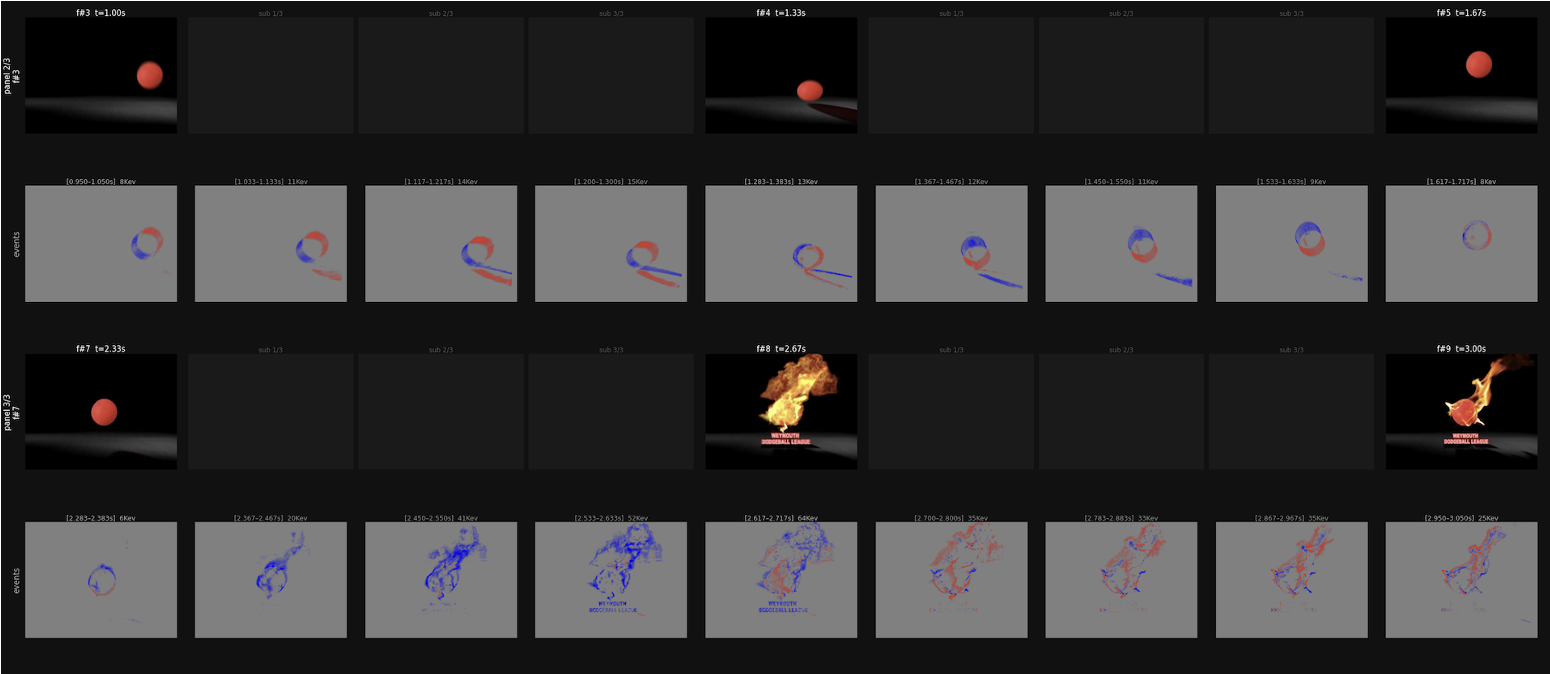}
  \caption{Converted ChatUniVi clip spanning 3\,s. \emph{Top row:} three
  consecutive source RGB frames. \emph{Bottom row:} converted event
  accumulations over 100\,ms windows covering the original and FILM-upsampled
  frames. Only moving regions generate events; the static background is
  silent.}
  \label{fig:conversion}
\end{figure}

\begin{figure}[htbp]
  \centering
  \includegraphics[width=\linewidth]{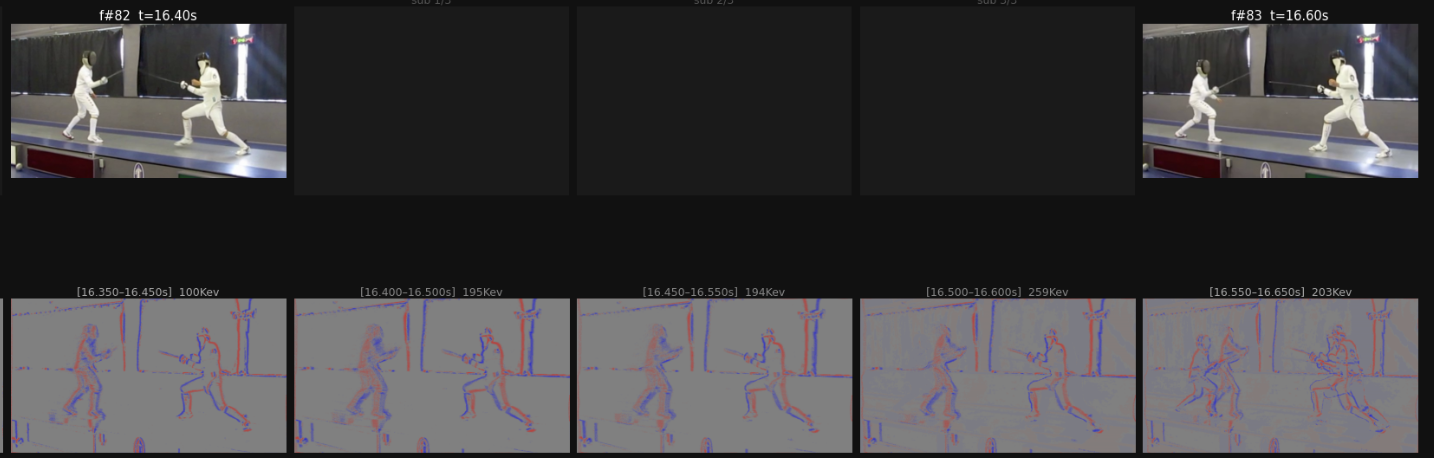}
  \caption{Converted VidSTG clip spanning 0.2\,s. \emph{Top row:} two
  consecutive source RGB frames. \emph{Bottom row:} converted
  event accumulations over 100\,ms windows covering the original and optical-flow-upsampled frames.}
  \label{fig:conversion_fig_vidstg}
\end{figure}

\begin{figure}[htbp]
  \centering
  \includegraphics[width=0.48\linewidth]{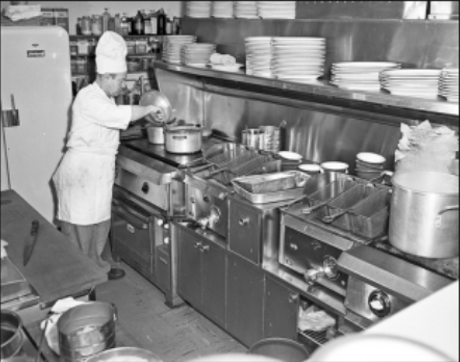}
  \hfill
  \includegraphics[width=0.48\linewidth]{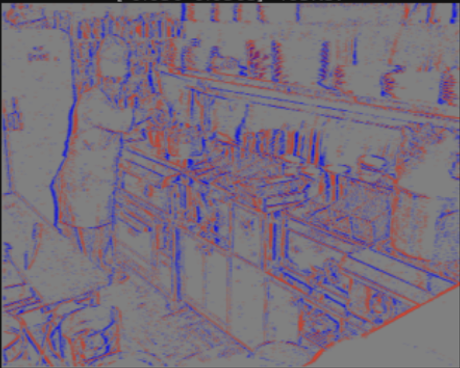}
  \caption{Saccade conversion of a still image. \emph{Left:} source RGB image
  from LLaVA-1.5. \emph{Right:} ON/OFF event accumulation after the
  $\pm 8$\,px saccade (ON red, OFF blue). The saccade fires events
  across the full image extent, providing training signal that a static image
  would otherwise not produce.}
  \label{fig:conversion_image_llava}
\end{figure}

\begin{figure}[htbp]
  \centering
  \includegraphics[width=\linewidth]{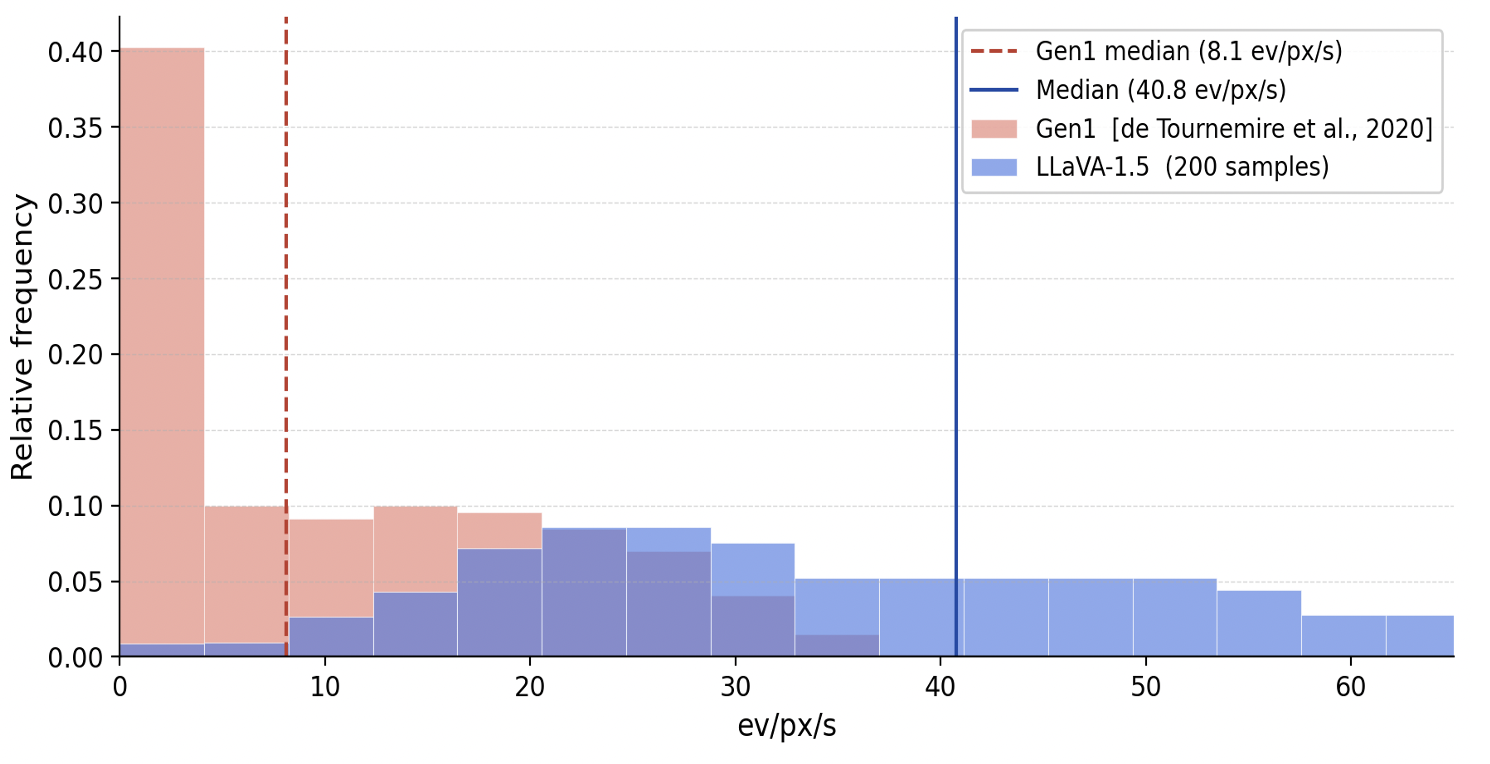}
  \caption{Distribution of events per pixel per second for saccade-converted
  LLaVA-1.5 images (200 images) and for real Gen1
  recordings~\citep{detournemire2020gen1}. Medians: LLaVA-1.5 40.8,
  Gen1 8.1\,\evpxs.
  }
  \label{fig:conversion_llava}
\end{figure}

\end{document}